# Unlocking Historical Clinical Trial Data with ALIGN: A Compositional Large Language Model System for Medical Coding

Nabeel Seedat[1,3]*, Caterina Tozzi[2], Andrea Hita Ardiaca[2],
Mihaela van der Schaar[1‡], James Weatherall [3‡], Adam Taylor [3‡]

[1] Department of Applied Mathematics and Theoretical Physics, University of Cambridge, Cambridge, United Kingdom

[2] Data Science & Artificial Intelligence, BioPharmaceuticals R&D, AstraZeneca, Barcelona, Spain

[3] Data Science & Artificial Intelligence, BioPharmaceuticals R&D, AstraZeneca, Cambridge, United Kingdom

* Work done at AstraZeneca

‡ Equal advising

**Corresponding authors:**

Nabeel Seedat
ns741@cam.ac.uk
University of Cambridge

Adam Taylor
adam.taylor2@astrazeneca.com
AstraZeneca

## ABSTRACT:

**Background:** The reuse of historical clinical trial data has significant potential to accelerate medical research and drug development. However, interoperability challenges, particularly with missing medical codes, hinder effective data integration across studies. While Large Language Models (LLMs) offer a promising solution for automated coding without labeled data, current approaches face challenges on complex coding tasks.

**Methods:** We introduce ALIGN, a novel compositional LLM-based system (i.e. compound AI system) for automated, zero-shot medical coding. ALIGN follows a three-step process: (1) diverse candidate code generation; (2) self-evaluation of codes and (3) confidence scoring and uncertainty estimation enabling human deferral to ensure reliability. We evaluate ALIGN on harmonizing medication terms into Anatomical Therapeutic Chemical (ATC) and medical history terms into Medical Dictionary for Regulatory Activities (MedDRA) codes extracted from 22 immunology trials (10 rheumatoid arthritis, 12 systemic lupus erythematosus) and compare to LLM baselines, including prompting and Retrieval Augmented Generation (RAG).

**Results:** ALIGN outperformed the LLM baselines, while also providing capabilities for trustworthy deployment. For MedDRA coding, ALIGN achieved high accuracy across all levels, matching RAG and excelling at the most specific levels (87-90% for HLGT). For ATC coding, ALIGN demonstrated superior performance, particularly at lower hierarchy levels (ATC Level 4), with 72-73% overall accuracy and 86-89% accuracy for common medications, outperforming baselines by 7-22%. ALIGN's uncertainty-based deferral improved accuracy by 17% to ±90% accuracy with 30% deferral, notably enhancing performance on uncommon medications. ALIGN achieves this cost-efficiently at $0.0007 and $0.02 per code for GPT-4o-mini and GPT-4o, reducing barriers to clinical adoption.

**Conclusion:** ALIGN advances automated medical coding for clinical trial data, contributing to enhanced data interoperability and reusability. Its robust performance and human-in-the-loop reliability positions ALIGN as a promising tool to unlock the potential of historical clinical trial data, improve clinical research and accelerate drug development.

# INTRODUCTION

Reusing historical clinical trial data offers transformative opportunities for medical research and drug development, offering the potential to reduce sample sizes, enhance statistical power and uncover broader medical insights obscured in individual datasets[1–3]. TransCelerate BioPharma's DataCelerate initiative [4] exemplifies this providing access to de-identified placebo and standard-of-care data[3]. Conservative estimates suggest such historical trial data reuse could reduce unnecessary recruitment by around 25%[1], reducing patient burden and speeding up new therapy delivery.

Despite this promise, a critical bottleneck persists: the lack of interoperability across diverse data sources [5–8]. This issue affects various healthcare settings, from large biopharmaceuticals[9] and hospital systems to small clinical trials and research networks. Consequently, the arduous manual harmonization effort required (3-6 months for dozens of trials), often forces clinical researchers to omit valuable cross-trial data or restrict analysis to easily harmonized subsets. Addressing this challenge is critical technically but also clinically, as it could unlock insights from cross-trial patient data previously hindered by dataset incompatibility.

In this context, standardized medical coding systems play are essential for interoperability. Examples include the Anatomical Therapeutic Chemical (ATC) classification[10] and the Medical Dictionary for Regulatory Activities (MedDRA) [11], which ensure consistent encoding and interpretation of data across studies, enabling diverse datasets to 'speak the same language' [6,7]

However, even with standard coding systems, clinical trials—often considered the gold standard of clinical data—are not immune to coding missingness, variability and inconsistencies. To illustrate these challenges, we examined immunology trial data from TransCelerate[3]. As shown in Figure 1 there is significant missingness of codes across different studies. Moreover, coding challenges are not limited to terminology differences; but also encompass structural, contextual and practical issues that arise within individual studies and when integrating multiple studies (see Table S1). These findings highlight the need for innovative solutions to enhance interoperability and unlock historical trial data's scientific potential.

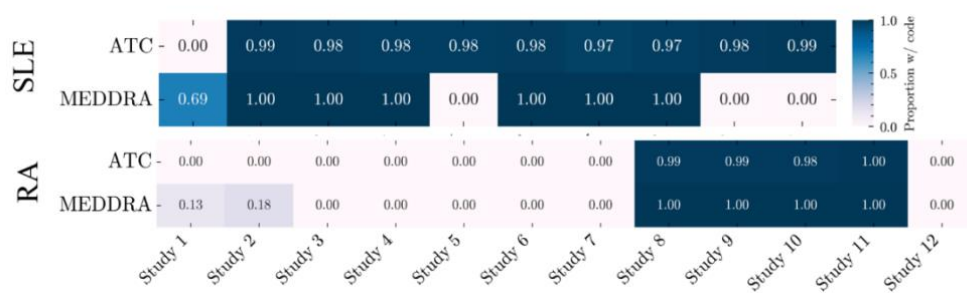

*Figure 1.* *Missingness of codes is pervasive across different trials: There is significant variability in the presence of standardized medical codes (ATC and MedDRA) across both Rheumatoid Arthritis (RA) and Systematic Lupus Erythematosus (SLE) studies. The missingness of codes in studies can be seen both within a therapeutic indication (RA vs SLR) and across indications, highlighting the need for automated medical coding solutions to enable interoperability of datasets.*

Recent advancements in Large Language Models (LLMs) have sparked interest to automate medical coding. This interest stems from LLMs' extensive 'prior knowledge' acquired through large-scale pretraining[12,13] and their capabilities for contextual understanding [14,15]. This offers the promise to perform tasks without labeled data (i.e. zero-shot), facilitating scalability across various clinical domains.

Prior studies have explored the use of LLMs for medical coding (ICD codes [16]), ranging from direct prompting [17–21], to more sophisticated Retrieval Augmented Generation (RAG) (also known as retrieve and rank methods)[22,23]. However, these methods treat medical coding as a single-step information extraction task, failing to capture the complex, multi-step reasoning process required for accurate coding. Consequently, they exhibit significant limitations, including poor code prediction performance and generating spurious codes. This highlights a critical gap between the current use of LLMs and the nuanced decision-making process employed by human coders, who consult coding manuals, consider context and verify their decisions.

To address these challenges, we propose ALIGN, a compositional LLM medical coding system (i.e., compound AI system[24]) that leverages the reasoning capabilities of LLMs while incorporating explicit verification. ALIGN uses a multi-step process where the LLM proposes codes, considers alternatives, verifies selections based on external grounded information and produces uncertainty estimates to enable human oversight —an essential consideration when LLM outputs are used in clinical settings. The consequent improvements to automated LLM-based medical coding enhances data interoperability in a reliable and time efficient manner, helping to unlock the full potential of historical trial data for both clinical and drug discovery applications --- ultimately contributing to improved patient outcomes.

# METHOD

**Study Design and Data Sources**

We developed **ALIGN** (**A**utomated **L**LM data Interoperability by **G**enerative medical codi**N**g), a compositional LLM system (i.e. compound AI system) designed for automated zero-shot medical coding. Our focus on zero-shot medical coding addresses the challenge of applying ALIGN to diverse data, without requiring new training data for each new scenario. We evaluated ALIGN on two medical coding systems widely used in clinical trials: ATC codes for concomitant medication names and MedDRA for medical history terms (also applicable to adverse events).

**Dataset construction**

We utilized historical placebo/standard of care data from 22 immunology clinical trials from TransCelerate BioPharma: 10 rheumatoid arthritis (RA) and 12 systemic lupus erythematosus (SLE) studies, comprising multiple trials, sponsors and years.

For ATC coding, we extracted medication names, dosages, and administration routes (when available) from CDISC concomitant medication files. We constructed our evaluation dataset from patient records with pre-existing ATC codes (level 4), annotated by medical coders. We identified unique combinations of medication names and routes of administration, as these can affect coding decisions, selecting the most commonly assigned ATC code for each unique pair to mitigate human variability.

For MedDRA coding, we used the medical history files, extracting the preferred term names as the query. Similar to ATC codes, we leverage records with pre-existing ground truth codes, grouping identical terms and selecting the most frequently assigned MedDRA codes (System Organ Class - SOC, Higher Level Group Term - HLGT, Higher Level Term - HLT).

Although tested on immunology data, ALIGN's zero-shot approach can be applied to other therapeutic areas. ALIGN's flexibility stems from its processing of fundamental clinical data elements – medication names, dosages, and medical histories – which are common across all therapeutic areas and other clinical datasets.

The evaluation datasets[*] are outlined below, with further details in Figure S1:

SLE dataset:

- ATC: 1,895 unique ATC codes, with a median code frequency of 2.0 (interquartile range: 1.0 to 7.0, maximum: 10,474).
- MedDRA: 26 unique SOC codes (median frequency: 293, IQR: 145 to 518, max: 1,940), 246 unique HLGT codes (median frequency: 12, IQR: 3 to 42, max: 653), and 655 unique HLT codes (median frequency: 4, IQR: 1 to 12, max: 568).

RA dataset:

- ATC: 2,019 unique ATC codes (median frequency: 2.0, IQR: 1.0 to 10.0, max: 8,063).
- MedDRA: 26 unique SOC codes (median frequency: 74, IQR: 44 to 127, max: 484), 186 unique HLGT codes (median frequency: 5, IQR: 2 to 13, max: 262), and 390 unique HLT codes (median frequency: 2, IQR: 1 to 5, max: 129).

---

[*] Note: Since we assess zero-shot performance, we use all the data as the test set.

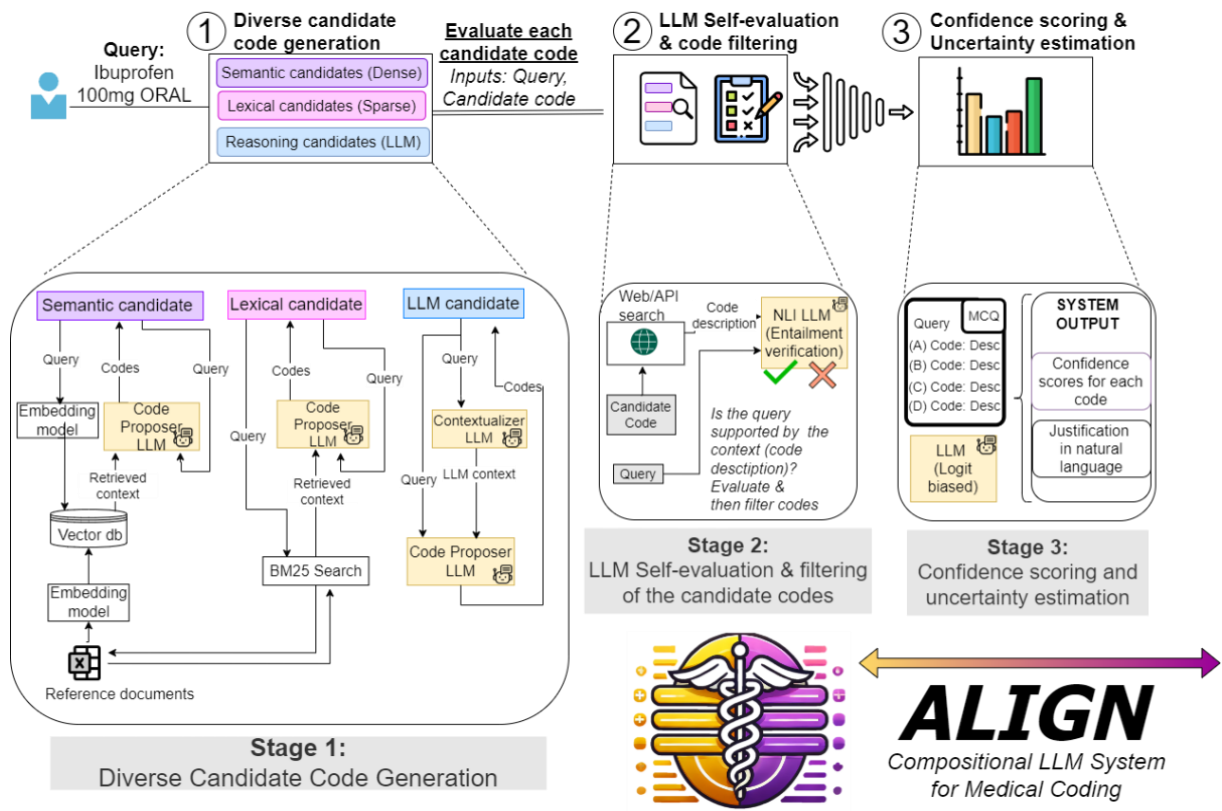

*Figure 2.* ALIGN is a compositional Large Language Model (LLM) system for medical coding. ALIGN takes a receives a query such as medication name or medical history and proceeds through three main stages: (1) Diverse candidate code generation, (2) LLM self-evaluation and code filtering to automatically filter spurious and inconsistent codes; and (3) Confidence scoring and uncertainty estimation, which provides a confidence score and permits human-in-the-loop deferral of uncertain cases, thereby ensuring reliability and trustworthiness.

**ALIGN System Architecture**

ALIGN's compositional LLM system (Figure 2) integrates several components designed to mimic the complex, multi-step reasoning process employed by human medical coders and has the following three stages:

(1) **Diverse Candidate Code Generation**

ALIGN employs three complementary approaches to generate diverse candidate codes: dense retrieval, BM25 retrieval and LLM reasoning.

**Dense Retrieval:** We use ChromaDB (version 0.5.1) as the vector store and OpenAI's text-embedding-3-large model to produce 1536-dimensional vectors. We index ATC and MedDRA codes along with their metadata (e.g. full term name, description, dosage etc.). During retrieval, input queries are embedded using the same model and the top-k (k=10) most similar candidate codes are retrieved based on cosine similarity.

**BM25 Retrieval**: We employ the Best Match 25 (BM25) algorithm for sparse retrieval using bm25s [25] (version 0.1.10) to index the descriptions of ATC and MedDRA codes. Sparse retrieval is performed by matching keywords from the query against the index. BM25 is widely used in information retrieval and web search, calculating relevance scores based on term frequency and inverse document frequency. We retrieve the top-k (k=10) candidate codes based on the relevance scores.

**LLM Reasoning:** We utilize an LLM to generate candidate codes as follows:

- Query Contextualization: A separate LLM generates contextual information about the queried medication or condition.
- Alternative Name Generation: The LLM generates synonyms for the query.
- Candidate Code Generation: LLM via chain-of-thought (CoT) [26] reasons over the enriched context to propose candidate codes.

### (2) LLM Self-Evaluation

To ensure code validity and mitigate spurious candidate codes, we implemented a self-evaluation mechanism that leverages external grounding. First, ALIGN retrieves grounded descriptions for each candidate code. For ATC codes, ALIGN queries and scrapes ATC/DDD[27] for the relevant descriptions. Similarly, for MedDRA terms, ALIGN retrieves standardized descriptions from the official MedDRA source [28]. These descriptions serve as contextual grounding information for the verification process.

ALIGN employs a separate LLM to perform a natural language inference (NLI) task—also known as textual entailment verification [29,30]—to evaluate whether the medication query or medical condition is logically entailed by or consistent with the retrieved context (code descriptions). This step provides external grounding for the candidate codes by ensuring they are validated against authoritative sources. We filter/prune those codes which do not logically entail as invalid or spurious --- mirroring how humans validate codes.

**Confidence Scoring and Uncertainty Estimation**

To provide robust uncertainty estimates and enable human deferral, we implemented a two-stage approach for confidence scoring and uncertainty estimation.

While language models may not be well-calibrated at the sequence level, recent research has shown LLM's provide good uncertainty estimates which are well calibrated at token-level [31–33] --- which is especially amenable to multiple choice question (MCQ)[32]. Consequently, to leverage this property we reformulate the code prediction task as a multiple choice question (MCQ), presenting the LLM with the candidate codes as MCQ options (e.g. (A), (B), (C) ). To account for scenarios where no viable candidates exist, we included a "None of the above" option, ensuring the LLM is not forced to select an inappropriate candidate code.

We used logit biasing to restrict the LLM's output to the presented single token options when prompting it to reason about and select the best code. Specifically, we increased the logit scores of the tokens corresponding to the candidate codes by a fixed value (in our case, 100), effectively constraining the LLM's choices to the presented options. We then extract the log-probabilities for each as confidence scores, resulting in a sorted prediction set. Subsequently, we apply a softmax function with temperature scaling to ensure the scores sum to 1.

Finally, we then estimate uncertainty by computing the predictive entropy of the prediction set, where higher entropy indicates greater uncertainty. This ensures ALIGN "knows what it doesn't know," flagging ambiguous cases for review. The threshold can be adjusted to balance automation with human oversight, essential for reliable and trustworthy ML in clinical settings.

***Conformal Prediction Extension:*** Going beyond our zero-shot setting, our system can easily be extended with conformal prediction [34] using a labeled calibration dataset. This provides prediction sets with frequentist statistical guarantees on coverage. Details are provided in the supplementary materials.

**Evaluation Metrics**

We evaluated coding performance primarily based on code prediction accuracy given our evaluation set consists of unique codes. Additionally, we assess ALIGN's prediction set coverage (ALIGN (Set)).

We assessed performance across multiple levels of the ATC and MedDRA hierarchies (i.e. varying levels of specificity), thereby accommodating various research needs in clinical and pharmacovigilance contexts.

- ATC codes: level 1 (anatomical), level 2 (therapeutic), level 3 (pharmacological), and level 4 (chemical).
- MedDRA terms: System Organ Class (SOC), High Level Group Term (HLGT), and High Level Term (HLT).

**Experimental Setup and Baseline Comparisons**

We assessed GPT-4o-mini and GPT-4o as LLM backbones. These models are internally compliant versions via Azure, aligning with our data usage agreement with TransCelerate and adhering to AstraZeneca's internal legal and GxP compliance requirements. All experiments used a temperature setting of 0.7 (as we found no sensitivity) with maximum tokens of 500.

We compared ALIGN against the following LLM baseline methods used in prior LLM coding works (applicable in the zero-shot setting):

- **LLM**: directly querying/prompting the LLM to generate the corresponding codes. Both vanilla prompting and Chain of Thought (CoT) [26] prompting were evaluated.

- **Retrieval-Augmented Generation (RAG)** [35] (also referred to as Retrieve-and-Rank): combines information retrieval of candidates and LLM-based ranking and is widely used for LLM tasks. We perform dense retrieval on a vector database to obtain candidate codes for a given query. The LLM is then prompted to rank and select the most appropriate option. By incorporating relevant retrieved information, this approach enhances the LLM's ability to generate accurate responses.

To account for the stochastic nature of LLMs, we repeated all experiments five times, reporting mean ± standard deviation. ALIGN and baselines were implemented using the DSPy framework [36].

## RESULTS

This section presents results for MedDRA and ATC coding tasks. We report results using GPT-4o-mini in the main manuscript as its cost-effectiveness makes it practical for large-scale, real-world applications. Results using GPT-4o are included in the supplementary section.

### MedDRA *coding performance evaluation.*

Table 1 shows that both ALIGN and RAG significantly outperform vanilla LLMs in coding medical history terms to MedDRA for both RA and SLE. As expected, accuracy decreases progressively from SOC to HLGT to HLT levels, reflecting the increasing granularity. That said, performance remains high even for the most specific HLT codes (ALIGN: 86.67% for RA, 90.22% for SLE; RAG: 87.15% for RA, 89.55% for SLE).

The comparable performance between RAG and ALIGN is expected in this context, given the predominantly 1:1 mapping of terms to codes. Specifically, this task primarily involves semantic mapping, without requiring complex reasoning, making it suitable for retrieval-based approaches.

Despite similar performance, we recommend ALIGN for its unique uncertainty quantification capability, allowing it to flag cases for human review, a key advantage over retrieval baselines.

*Table 1.* MedDRA code prediction accuracy (mean±std) for the three levels of the MedDRA hierarchy: System Organ Class (SOC), Higher Level Group Term (HLGT) and Higher Level Term (HLT) using GPT-4o-mini as the backbone LLM. RAG-based baselines and ALIGN outperform vanilla LLM prompting approaches and they both have relatively similar performance across the three MedDRA levels given the mapping of query-code is largely semantic.

|  | RA | | | SLE | | |
| --- | --- | --- | --- | --- | --- | --- |
|  | SOC | HLGT | HLT | SOC | HLGT | HLT |
| **LLM** | 61.86±3.65 | 60.60±3.60 | 55.98±2.62 | 58.76±2.02 | 56.75±2.22 | 52.97±1.80 |
| **LLM (CoT)** | 62.10±3.17 | 61.08±3.28 | 56.76±4.39 | 60.01±3.34 | 58.26±3.42 | 55.28±2.98 |
| **RAG** | 96.52±0.65 | 95.44±1.11 | 87.15±1.72 | 96.60±2.91 | 94.28±3.11 | 89.55±3.07 |
| **RAG (CoT)** | 95.98±3.57 | 94.53±4.50 | 85.89±5.57 | 97.36±2.54 | 95.27±2.61 | 90.54±2.53 |
| **ALIGN** | 96.22±0.92 | 94.95±0.84 | 86.67±0.73 | 97.70±0.18 | 95.23±0.15 | 90.22±0.20 |
| **ALIGN (Set)** | 97.31±0.82 | 96.18±0.91 | 88.12±0.81 | 97.41±0.25 | 95.22±0.27 | 90.49±0.22 |

## ATC coding performance evaluation.

The ATC coding results in Table 2 (a) highlight the complexities of ATC coding compared to MedDRA. While retrieve-rank methods outperform vanilla LLMs, ALIGN improves upon retrieve-rank approaches, particularly at lower levels of the ATC hierarchy (level 3 and 4).

This performance gap underscores the complexity of ATC coding, which often involves 1-to-many mappings and requires reasoning about medication context, administration, and dosage to assign the correct code. ALIGN's superior performance at lower ATC levels demonstrates its capability to handle nuanced reasoning beyond semantic matching.

We further analyzed performance on common and uncommon medications (based on the Pareto principle), with common medications making up 80% of occurrences and uncommon the remainder.

Table 2(b) shows **ALIGN's** particularly strong performance on common medications, with **±10% performance improvement** at ATC level 4 compared to the overall results. This improvement has real-world implications, ensuring more reliable coding for the majority of cases in clinical trials and practice, which not only saves time but also reduces inconsistencies, providing faster and more reliable cross-trial data integration.

Conversely, the relatively lower performance on uncommon medications (Table 2 (c)) highlights challenges to code rarer drugs. This can be attributed to several factors, including limited contextual information for uncommon medications in the LLMs training corpus, the complexity of rare cases that might involve specialized treatments or unique administration protocols and the potential for specific drugs that don't fit existing ATC categories (e.g. Epikajutsuto, a traditional Japanese medication). Nevertheless, these findings underscore the value of ALIGN's approach in improving coding accuracy for the most impactful subset of medications while also indicating areas for future improvement in handling uncommon drugs.

Additionally, we highlight the cost efficiency of ALIGN, which is relevant for large-scale real-world deployments, with a cost efficiency of $0.0007 and $0.02 per code for GPT-4o-mini and GPT-4o LLM backbones respectively (see Figure S2) ---thereby lowering the barrier to adoption in clinical settings.

*Table 2*. ATC code prediction accuracy (mean±std) using GPT-4o-mini as the backbone LLM. We report (a) overall results for all unique query-administration pairs, (b) common codes representing 80% of all occurrences and (c) uncommon codes representing the remaining 20%. ALIGN outperforms the baselines especially on the lower ATC levels and exhibits strong performance on the most commonly used concomitant medications.

| | RA | | | | | SLE | | | |
|---|---|---|---|---|---|---|---|---|---|
| **(a) Overall** | ATC Level 1 | ATC Level 2 | ATC Level 3 | ATC Level 4 | **(a) Overall** | ATC Level 1 | ATC Level 2 | ATC Level 3 | ATC Level 4 |
| **LLM** | 85.16 ± 0.42 | 76.56 ± 0.21 | 65.72 ± 0.58 | 49.99 ± 0.78 | **LLM** | 84.88 ± 0.41 | 74.81 ± 0.37 | 65.71 ± 0.46 | 48.91 ± 0.46 |
| **LLM (CoT)** | 87.60 ± 0.35 | 80.63 ± 0.52 | 71.02 ± 0.69 | 50.84 ± 0.91 | **LLM (CoT)** | 86.34 ± 0.13 | 78.75 ± 0.40 | 69.94 ± 0.69 | 49.93 ± 0.49 |
| **RAG** | 81.83 ± 0.12 | 76.43 ± 0.15 | 72.24 ± 0.12 | 65.27 ± 0.11 | **RAG** | 82.04 ± 0.33 | 76.44 ± 0.36 | 71.73 ± 0.20 | 66.29 ± 0.20 |
| **RAG (CoT)** | 81.88 ± 0.36 | 76.47 ± 0.26 | 71.89 ± 0.28 | 65.07 ± 0.26 | **RAG (CoT)** | 81.89 ± 0.34 | 76.31 ± 0.43 | 71.52 ± 0.43 | 65.74 ± 0.46 |
| **ALIGN** | 85.92 ± 0.25 | 80.38 ± 0.14 | 77.60 ± 0.13 | 72.20 ± 0.33 | **ALIGN** | 86.58 ± 0.26 | 81.29 ± 0.29 | 78.31 ± 0.26 | 73.33 ± 0.38 |
| **ALIGN (Set)** | 93.76 ± 0.22 | 88.93 ± 0.28 | 84.57 ± 0.29 | 79.18 ± 0.30 | **ALIGN (Set)** | 91.96 ± 0.50 | 87.05 ± 0.42 | 82.91 ± 0.34 | 77.92 ± 0.61 |
| **(b) Common** | | | | | **(b) Common** | | | | |
| **LLM** | 95.84 ± 0.40 | 92.08 ± 0.63 | 82.97 ± 1.15 | 66.53 ± 2.96 | **LLM** | 98.33 ± 0.00 | 89.67 ± 1.63 | 79.33 ± 2.00 | 65.00 ± 3.80 |
| **LLM (CoT)** | 96.44 ± 0.79 | 93.86 ± 0.97 | 88.51 ± 1.01 | 69.70 ± 3.84 | **LLM (CoT)** | 96.67 ± 1.83 | 90.33 ± 1.25 | 84.67 ± 3.06 | 66.33 ± 5.31 |
| **RAG** | 92.48 ± 0.79 | 91.09 ± 0.63 | 90.89 ± 0.74 | 85.35 ± 0.97 | **RAG** | 94.00 ± 0.82 | 89.00 ± 0.82 | 87.33 ± 0.82 | 76.67 ± 1.0 |
| **RAG (CoT)** | 91.49 ± 1.01 | 90.30 ± 1.15 | 90.30 ± 1.15 | 84.36 ± 0.74 | **RAG (CoT)** | 93.67 ± 0.67 | 88.67 ± 0.67 | 87.00 ± 0.67 | 77.00 ± 0.67 |
| **ALIGN** | 95.59 ± 1.38 | 93.16 ± 1.28 | 94.48 ± 1.23 | 88.96 ± 1.44 | **ALIGN** | 96.87 ± 0.70 | 94.79 ± 0.07 | 94.80 ± 2.42 | 86.11 ± 2.45 |
| **ALIGN (Set)** | 98.02 ± 1.08 | 97.14 ± 1.64 | 96.03 ± 0.87 | 91.39 ± 0.46 | **ALIGN (Set)** | 100.00 ± 0.00 | 99.31 ± 0.84 | 97.23 ± 0.83 | 92.72 ± 1.23 |
| **(c) Uncommon** | | | | | **(c) Uncommon** | | | | |
| **LLM** | 84.08 ± 0.47 | 75.00 ± 0.25 | 63.98 ± 0.61 | 48.33 ± 1.03 | **LLM** | 84.25 ± 0.43 | 74.11 ± 0.36 | 65.08 ± 0.48 | 48.16 ± 0.59 |
| **LLM (CoT)** | 86.71 ± 0.36 | 79.30 ± 0.51 | 69.26 ± 0.71 | 48.94 ± 0.71 | **LLM (CoT)** | 85.86 ± 0.06 | 78.21 ± 0.39 | 69.25 ± 0.62 | 49.16 ± 0.44 |
| **RAG** | 80.76 ± 0.17 | 74.96 ± 0.18 | 70.36 ± 0.15 | 63.25 ± 0.06 | **RAG** | 81.48 ± 0.33 | 75.85 ± 0.35 | 71.00 ± 0.20 | 65.80 ± 0.20 |
| **RAG (CoT)** | 80.92 ± 0.42 | 75.08 ± 0.33 | 70.04 ± 0.28 | 63.13 ± 0.25 | **RAG (CoT)** | 81.34 ± 0.34 | 75.73 ± 0.43 | 70.79 ± 0.44 | 65.22 ± 0.47 |
| **ALIGN** | 86.59 ± 0.46 | 80.54 ± 0.68 | 76.52 ± 0.55 | 70.00 ± 0.47 | **ALIGN** | 86.10 ± 0.28 | 80.66 ± 0.31 | 77.57 ± 0.20 | 72.77 ± 0.48 |
| **ALIGN (Set)** | 93.44 ± 0.15 | 88.01 ± 0.64 | 82.81 ± 0.51 | 76.72 ± 0.79 | **ALIGN (Set)** | 91.97 ± 0.46 | 87.12 ± 0.39 | 83.01 ± 0.37 | 78.22 ± 0.65 |

**Uncertainty based human-in-the-loop integration improves coding reliability**

Despite ALIGN's improvement over baselines, the inherent complexities of ATC coding necessitate further improvements for reliable usage, particularly at lower ATC levels. ALIGN's approach to uncertainty quantification enables principled deferral to human experts (human-in-the-loop), a critical feature to enhance accuracy and trustworthiness in clinical settings.

Assuming correct human annotation for these deferred cases, we quantified final accuracy of the mapping with respect to simulated manual effort required. Figure 3 illustrates the impact of deferral on accuracy for both common and uncommon codes across RA and SLE datasets. Our uncertainty-based deferral outperforms random deferral, showing ALIGN's ability to identify cases it might misclassify—effectively "knowing what it doesn't know." The simulations of different percentages of human intervention reveal significant performance gains, particularly for uncommon codes. With approximately **30%** deferral, we observed accuracy improvements approaching **90%**, representing a substantial increase of over **15%** for rare medications. This difference in deferral impact highlights ALIGN's ability to ensure accuracy for routine cases while effectively using human expertise for challenging cases.

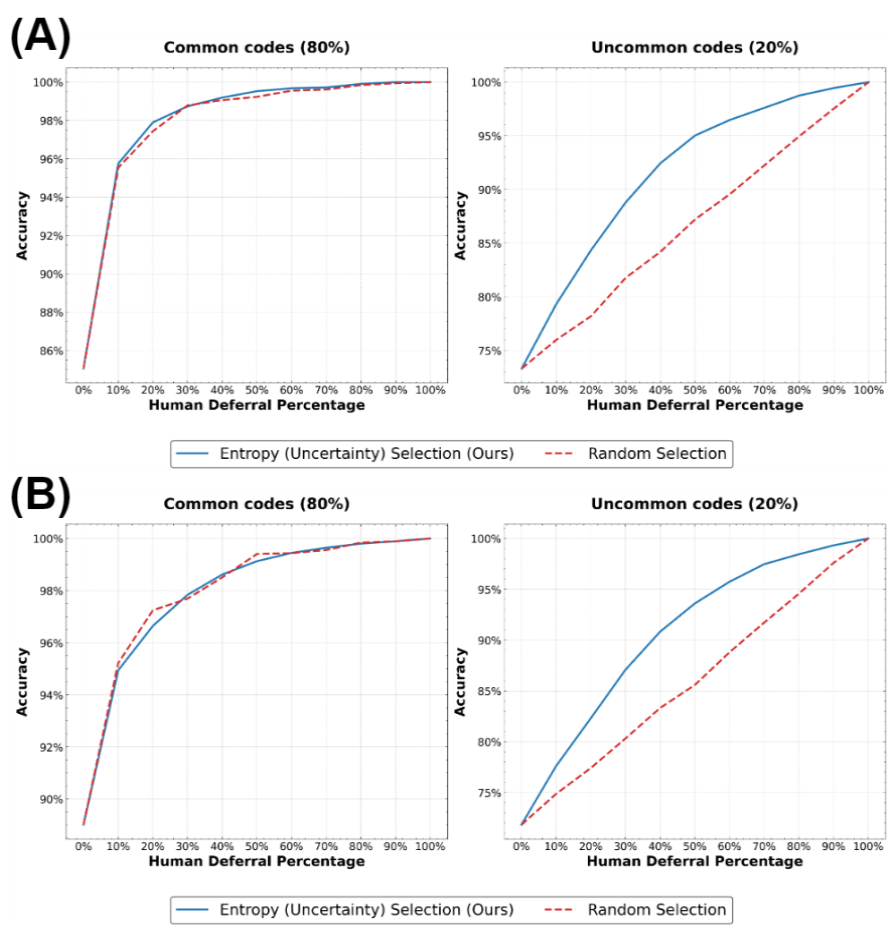

**Figure 3.** Simulating ATC coding performance for various human deferral percentages for ALIGN (GPT-4o-mini backbone) for both (Panel A) SLE and (Panel B) RA datasets. We see that human-in-the-loop deferral of code predictions leads to significant performance improvements --- especially on uncommon codes. Our selection mechanism using entropy for deferral outperforms random deferral highlighting the capability to capture uncertainty correctly (i.e. pinpoint those instances which would most benefit from human deferral).

# DISCUSSION

This study addresses the interoperability challenge in historical trial data reuse through a novel approach to standardize medical coding. Specifically, we introduce ALIGN, a compositional LLM system (or compound AI system) for zero-shot medical coding. ALIGN demonstrates superior performance over existing LLM-based methods in ATC and MedDRA coding. This advancement benefits drug development and clinical research by shortening harmonization (3-6 months for dozens of trials), speeding up timelines and reducing coding inconsistencies for more reliable clinical data. Additionally, ALIGN improves interoperability which can enable larger sample sizes, faster trials and reduced patient recruitment, contributing to accelerated medical research. Furthermore, ALIGN's zero-shot adaptability to diverse domains positions it as a valuable tool for both large-scale multi-center studies and smaller clinical studies, leveling the playing field in terms of data quality and interoperability --- potentially helping to bring new therapies to patients faster and enable more agile research in diverse clinical domains.

Technically ALIGN's self-evaluation mechanism excels in complex tasks like ATC coding, significantly reducing spurious or unreliable codes. Its uncertainty quantification capability enables human-in-the-loop intervention --- vital for trustworthy and responsible deployment of LLMs in healthcare. By selectively integrating human expertise for uncertain cases, we improve performance and reliability, reduce expert burden and enable more efficient AI-human collaboration.

Despite promising results, ALIGN has limitations, especially with uncommon codes, highlighting the need for human oversight. Future work should improve performance in these areas and expand evaluations to more medical indications, coding systems and other LLMs.

In conclusion, ALIGN's compositional LLM system represents an advancement in zero-shot automated medical coding for clinical trial data, contributing towards enhancing data interoperability and reusability. This advancement holds considerable potential for improving clinical analytics, accelerating drug development, reducing costs and ultimately improving patient outcomes. As the volume of trial data grows, systems like ALIGN will be key to unlocking the full potential of historical data and accelerating medical innovation.

## Acknowledgements:

The authors wish to thank the following companies who have contributed data to TransCelerate's DataCelerate platform, which facilitates the sharing of control arm data between participating member companies: AbbVie, Amgen, Astellas, AstraZeneca, Boehringer Ingelheim, Bristol-Myers Squibb, EMD Serono, Genentech, Janssen, Lilly, Novartis, Novo Nordisk, Pfizer, Sanofi, Shionogi, and UCB Biosciences. Neither TransCelerate Biopharma nor any of the above-referenced companies (excluding AstraZeneca) have contributed to, approved or are in any way responsible for the author's research results.

# Supplementary: Table of Contents

# Challenges of medical coding

*Table S1. Breakdown of the different challenging dimensions of medical coding as found for ATC codes from the Transcelerate RA and SLE datasets. The different dimensions highlight that the challenges go beyond semantics alone and even standardized data like clinical trials are not immune from interoperability challenges.*

| | Dimension | Description (and examples from Transcelerate data) |
|---|---|---|
| 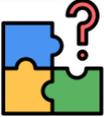 | Missing codes | Standardized codes are missing within or between studies. |
| 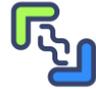 | Semantic Alignment | Ensuring consistent encoding and meaning of terms across different datasets.<br><br>*e.g. Different brand names referring to the same medication: Advil (US) vs Ibuprofen (generic) - M01AE01* |
| 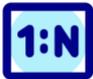 | One-to-Many Relationships in codes | Codes that have one-to-many mappings, varying based on factors like indication, route, and dosage.<br><br>*e.g. Acetylcysteine assigned different ATC codes (R05CB01, S01XA08, V03AB23) based on therapeutic use, administration route or dosage.* |
| 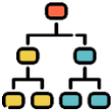 | Hierarchical Complexity | Multi-level coding structures<br><br>*e.g. Metformin: A10BA02 (biguanides, single drug type used to lower blood glucose levels) vs. A10BD (combinations with other oral blood glucose lowering drugs)* |
| 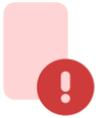 | Non-Adherence to Coding Standards | Inconsistent application of coding standards within and between studies, often due to off-label usage or lack of standardization protocols.<br><br>*e.g. Use of non-standard codes or off-label indications not properly captured by ATC, such as gynecological use of drugs like Methotrexate coded as G02CX (other gynae drugs) instead of L04AX/L01BA.* |
| 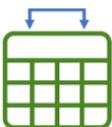 | Intra-Study Variability | Inconsistent coding within a single study<br><br>*e.g. Anovlar coded as G03AA, G03AB, or G03FA in the same study for the same admin route (human variability)* |
| 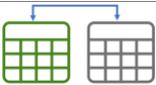 | Inter-Study Variability | Inconsistent coding between studies<br><br>*e.g. Anovlar coded differently across studies: study 1 (G03FA) vs. study 2 (G03AA)* |
| 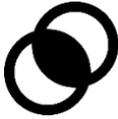 | Combination Products | Coding drugs that combine multiple agents<br><br>*e.g. Fixed-dose combination of metformin and sitagliptin: A10BD07* |
| 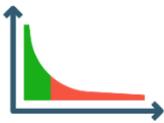 | Long Tail codes | Rare, herbal, or non-standard medications<br><br>*e.g. Goshakusan or Epikajutsuto (traditional Japanese medicines without standard ATC codes)* |

# Additional Dataset Details

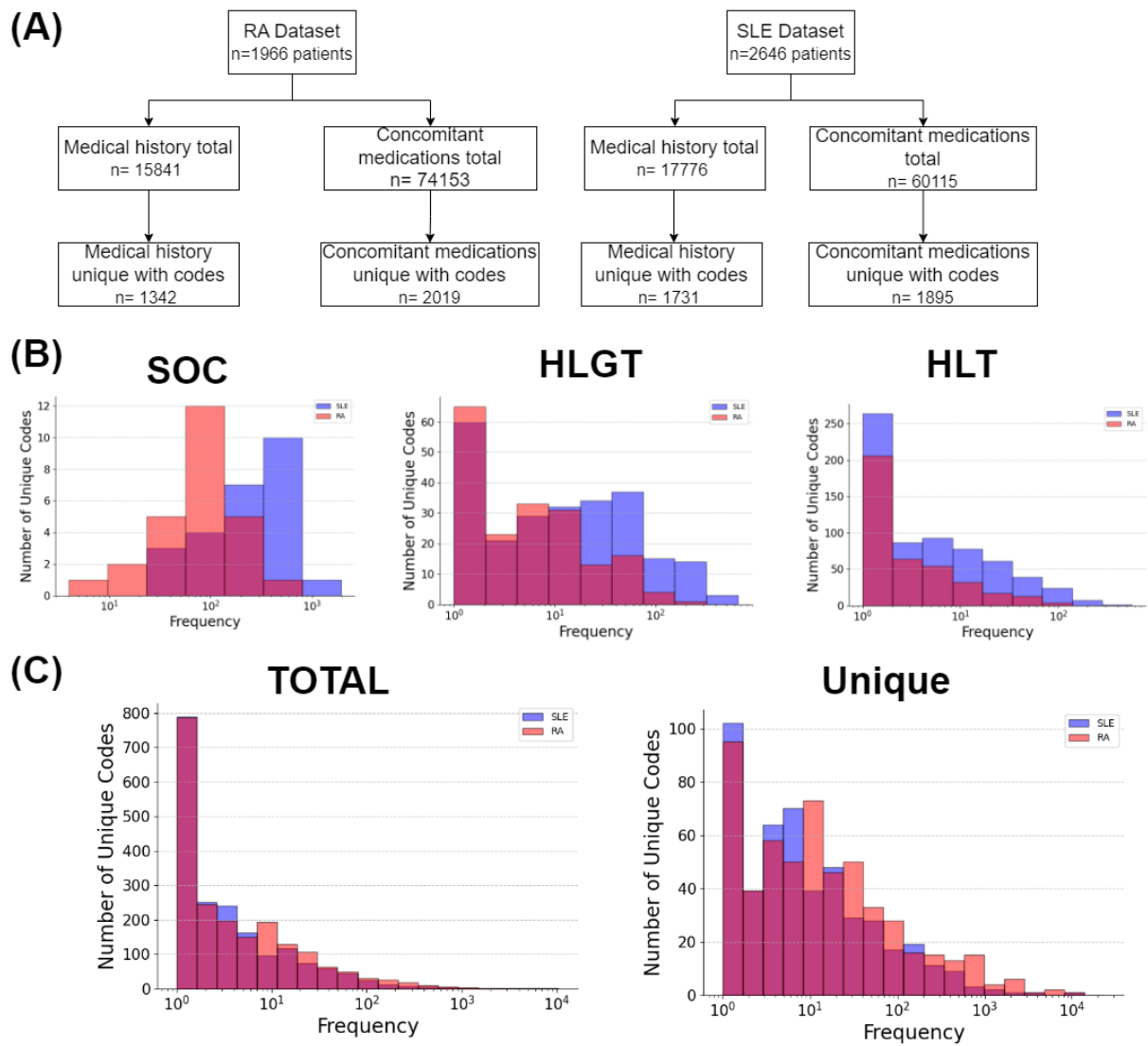

*Figure S1.* Panel A. Profile of the RA and SLE datasets illustrating the number of input entries and unique categories for both medical history and concomitant medication. The unique queries of medical history and concomitant medication-administration route pairs are derived from those studies capturing pre-existing annotations from medical coders. Panel B. Distributions of the three levels of MedDRA codes reflecting different levels of granularity – SOC, HLGT and HLT for both RA and SLE datasets. Panel C. Distributions of ATC codes both total and unique for both RA and SLE datasets

## Cost vs Performance Analysis

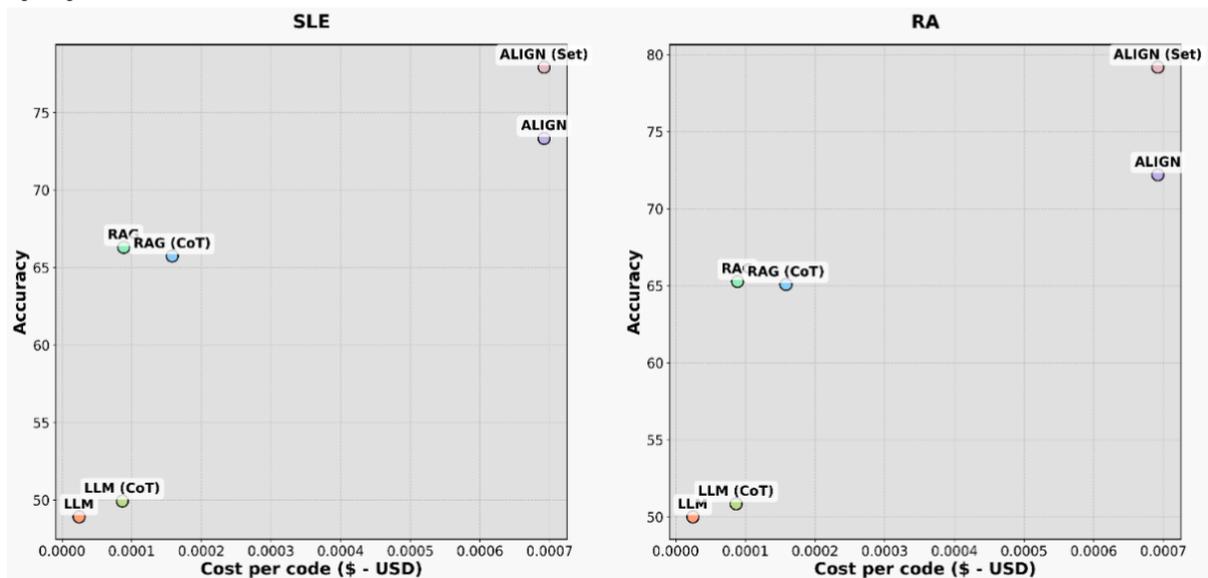

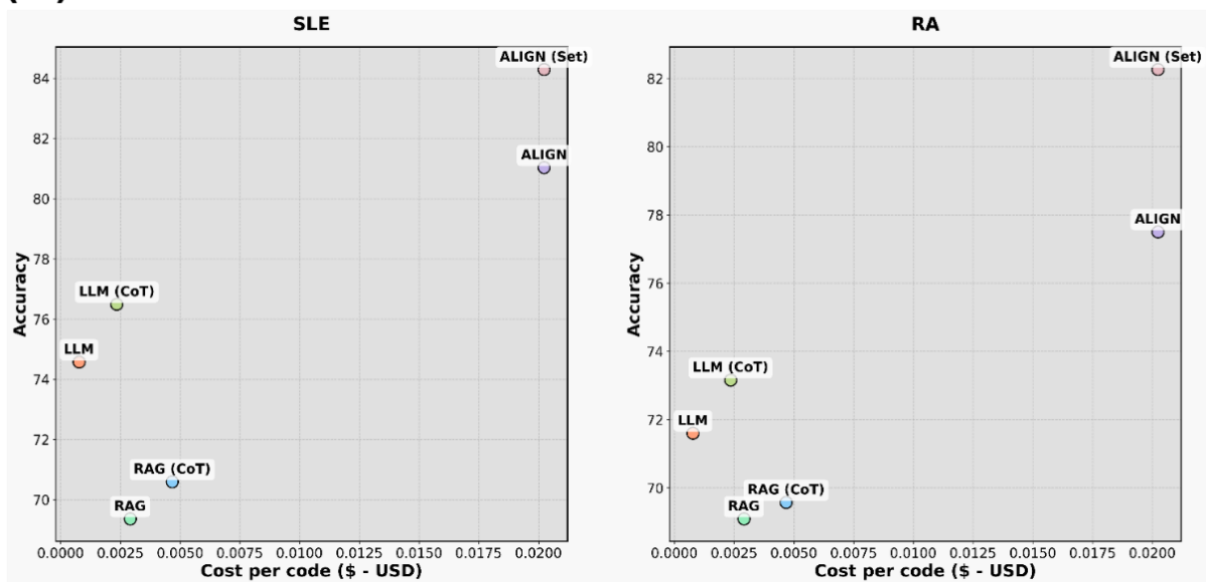

*Figure S2.* Cost vs Coding Performance trade-off with the following LLM backbones: (Panel A) GPT-4o-mini and (Panel B) GPT-4o. While ALIGN increases cost, it also achieves superior performance. Beyond performance the increased cost of ALIGN is due the capability for self-evaluation and confidence scoring/uncertainty estimation. The additional token counts are vital to ensure reliable performance, estimate system uncertainty and facilitate human intervention. Such capability is absent from baselines. Furthermore, there is a significant cost increase to using GPT-4o as the backbone vs GPT-4o-mini, especially if used at scale. This highlights in real-world settings at scale GPT-4o-mini would likely be preferred from a cost perspective.

# Additional Results: GPT-4o

*Table S2.* MedDRA code prediction accuracy (mean+-std) for the three levels of the MedDRA hierarchy: System Organ Class (SOC), Higher Level Group Term (HLGT) and Higher Level Term (HLT) using GPT-4o as the backbone LLM. RAG-based baselines and ALIGN outperform vanilla LLM prompting approaches and they both have relatively similar performance across the three MedDRA levels given the mapping of query-code is largely semantic.

|  | RA | | | SLE | | |
| --- | --- | --- | --- | --- | --- | --- |
|  | SOC | HLT | HLGT | SOC | HLT | HLGT |
| **LLM** | 72.55±8.14 | 71.71±8.16 | 65.65±7.16 | 70.56±8.23 | 68.85±8.42 | 65.05±8.05 |
| **LLM (CoT)** | 72.31±5.89 | 71.71±5.89 | 65.53±4.64 | 69.44±5.08 | 67.66±5.05 | 63.93±4.71 |
| **RAG** | 97.84±0.48 | 96.94±0.48 | 89.13±0.48 | 97.37±0.41 | 95.21±0.34 | 90.29±0.35 |
| **RAG (CoT)** | 98.98±0.15 | 97.96±0.22 | 90.09±0.19 | 98.91±0.15 | 96.85±0.16 | 92.15±0.08 |
| **ALIGN** | 96.50±1.13 | 95.50±0.85 | 87.19±0.71 | 95.05±0.27 | 93.18±0.24 | 88.28±0.24 |
| **ALIGN (Set)** | 98.06±0.40 | 97.02±0.38 | 89.06±0.23 | 97.75±0.20 | 95.85±0.17 | 90.97±0.20 |

**Table S3.** ATC code prediction accuracy (mean+-std) using GPT-4o as the backbone LLM. We report (a) overall results for all unique query-administration pairs, (b) common codes representing 80% of all occurrences and (c) uncommon codes representing the remaining 20%. ALIGN outperforms the baselines especially on the lower ATC levels and exhibits strong performance on the most commonly used concomitant medications.

| | RA | | | | SLE | | | | |
|---|---|---|---|---|---|---|---|---|---|
| (a) Overall | ATC Level 1 | ATC Level 2 | ATC Level 3 | ATC Level 4 | (a) Overall | ATC Level 1 | ATC Level 2 | ATC Level 3 | ATC Level 4 |
| LLM | 88.76 ± 0.38 | 83.78 ± 0.50 | 77.76 ± 0.34 | 71.60 ± 0.47 | LLM | 89.54 ± 0.33 | 83.88 ± 0.32 | 79.40 ± 0.47 | 74.57 ± 0.52 |
| LLM (CoT) | 90.32 ± 0.24 | 85.65 ± 0.31 | 80.00 ± 0.44 | 73.16 ± 0.35 | LLM (CoT) | 90.53 ± 0.32 | 85.92 ± 0.40 | 81.39 ± 0.54 | 76.49 ± 0.69 |
| RAG | 84.94 ± 0.42 | 79.88 ± 0.43 | 75.32 ± 0.49 | 69.09 ± 0.63 | RAG | 84.12 ± 0.17 | 78.59 ± 0.22 | 74.02 ± 0.28 | 69.36 ± 0.31 |
| RAG (CoT) | 84.87 ± 0.22 | 80.05 ± 0.42 | 75.67 ± 0.66 | 69.57 ± 0.54 | RAG (CoT) | 84.70 ± 0.42 | 79.42 ± 0.22 | 74.98 ± 0.29 | 70.59 ± 0.23 |
| ALIGN | 88.22 ± 0.36 | 83.93 ± 0.13 | 82.72 ± 0.53 | 77.50 ± 0.58 | ALIGN | 88.30 ± 0.28 | 84.50 ± 0.13 | 84.16 ± 0.19 | 81.03 ± 0.14 |
| ALIGN (Set) | 94.45 ± 0.33 | 91.38 ± 0.35 | 87.65 ± 0.48 | 82.26 ± 0.55 | ALIGN (Set) | 94.66 ± 0.16 | 90.25 ± 0.10 | 87.23 ± 0.21 | 84.29 ± 0.20 |
| (b) Common | | | | | (b) Common | | | | |
| LLM | 96.44 ± 0.49 | 94.06 ± 0.63 | 92.08 ± 0.63 | 89.70 ± 0.49 | LLM | 94.67 ± 0.67 | 89.67 ± 1.63 | 88.00 ± 1.63 | 82.33 ± 2.26 |
| LLM (CoT) | 96.44 ± 0.79 | 95.05 ± 0.63 | 93.47 ± 0.79 | 88.71 ± 0.49 | LLM (CoT) | 97.33 ± 1.33 | 92.00 ± 1.63 | 90.33 ± 1.63 | 82.67 ± 1.70 |
| RAG | 94.26 ± 0.40 | 92.28 ± 0.40 | 92.28 ± 0.40 | 87.72 ± 0.79 | RAG | 96.00 ± 0.82 | 89.33 ± 1.33 | 89.33 ± 1.33 | 79.00 ± 0.82 |
| RAG (CoT) | 93.47 ± 0.49 | 92.08 ± 0.00 | 92.08 ± 0.00 | 88.71 ± 0.79 | RAG (CoT) | 92.33 ± 1.70 | 86.00 ± 2.26 | 86.00 ± 2.26 | 80.00 ± 1.83 |
| ALIGN | 94.89 ± 0.57 | 92.66 ± 0.93 | 95.11 ± 0.89 | 88.89 ± 0.98 | ALIGN | 98.22 ± 0.06 | 93.93 ± 1.00 | 98.22 ± 0.06 | 90.04 ± 0.74 |
| ALIGN (Set) | 97.78 ± 0.03 | 96.44 ± 0.45 | 96.44 ± 0.45 | 90.22 ± 1.07 | ALIGN (Set) | 100.00 ± 0.00 | 98.22 ± 0.06 | 98.22 ± 0.06 | 91.82 ± 1.74 |
| (c) Uncommon | | | | | (c) Uncommon | | | | |
| LLM | 87.99 ± 0.39 | 82.75 ± 0.49 | 76.31 ± 0.31 | 69.78 ± 0.49 | LLM | 89.30 ± 0.33 | 83.61 ± 0.32 | 79.00 ± 0.49 | 74.21 ± 0.52 |
| LLM (CoT) | 89.70 ± 0.23 | 84.70 ± 0.32 | 78.65 ± 0.46 | 71.59 ± 0.35 | LLM (CoT) | 90.21 ± 0.35 | 85.63 ± 0.39 | 80.97 ± 0.55 | 76.20 ± 0.70 |
| RAG | 84.00 ± 0.48 | 78.64 ± 0.48 | 73.61 ± 0.53 | 67.22 ± 0.68 | RAG | 83.57 ± 0.18 | 78.08 ± 0.25 | 73.30 ± 0.29 | 68.91 ± 0.33 |
| RAG (CoT) | 84.00 ± 0.26 | 78.84 ± 0.46 | 74.02 ± 0.72 | 67.65 ± 0.65 | RAG (CoT) | 84.34 ± 0.38 | 79.11 ± 0.13 | 74.47 ± 0.21 | 70.15 ± 0.19 |
| ALIGN | 87.54 ± 0.42 | 83.04 ± 0.17 | 81.45 ± 0.50 | 76.33 ± 0.57 | ALIGN | 88.24 ± 0.70 | 84.16 ± 0.47 | 83.27 ± 0.39 | 80.27 ± 0.53 |
| ALIGN (Set) | 94.16 ± 0.19 | 90.95 ± 0.34 | 86.52 ± 0.64 | 81.34 ± 0.89 | ALIGN (Set) | 95.21 ± 0.28 | 90.93 ± 0.38 | 87.86 ± 0.54 | 84.77 ± 0.48 |

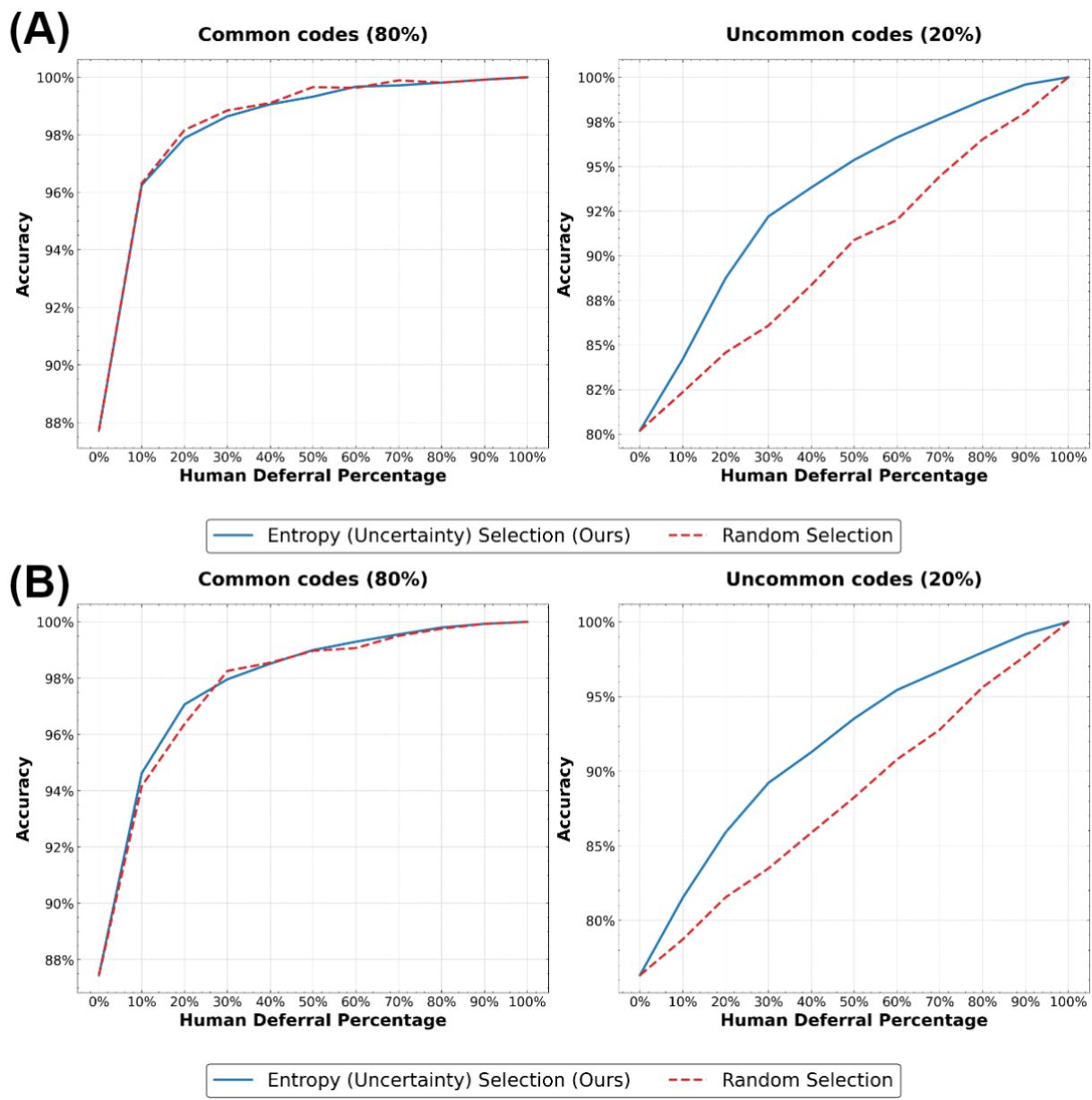

*Figure S3.* Simulating ATC coding performance for various human deferral percentages for ALIGN (GPT-4o backbone). Results are provided for SLE (Panel A) and RA (Panel B). We see that human-in-the-loop deferral of code predictions leads to a significant increase in performance --- especially on uncommon codes. Our selection mechanism using entropy for deferral outperforms random deferral highlighting the capability to capture uncertainty correctly (i.e. pinpoint those instances which would most benefit from human deferral).

# Uncertainty extension: Conformal Prediction

**Background:** Conformal Prediction (CP) [34] is a statistical framework that allows models to generate prediction sets with frequentist guarantees on user-defined coverage, without making any distributional assumptions. CP ensures that the true label is contained in the prediction set with a specified coverage probability (e.g., 95%).

Hence, given a new query test instance $x_{\text{test}}$ the model produces a set of potential predictions. CP then leverages a *calibration dataset* — a subset of labeled data (drawn in exchangeable manner from the population) — to calculate *nonconformity scores* that measure how unusual or unexpected the prediction for a particular instance is compared to the calibration data. The calibration step ensures that we can control the level of uncertainty in the prediction and provide frequentist guarantees on the correctness of the prediction set. This is useful in medical coding where reliability is paramount.

In Conformal Prediction, the coverage guarantee is formulated as:

$$P(y_{\text{true}} \in C(x_{\text{test}})) \geq 1 - \alpha$$

where $y_{\text{true}}$ is the true label, $C(x_{\text{test}})$ is the prediction set, and α is the significance level chosen by the user (e.g., α=0.05 for a 95% coverage guarantee). This ensures that, with high probability, the prediction set contains the true label.

**ALIGN extended with CP:** In scenarios where labeled data is available, we can integrate CP into our uncertainty estimation step of ALIGN. The steps for incorporating CP are as follows:

1. **Nonconformity Scoring**: For each instance in the calibration set, we calculate the nonconformity score based on the model's confidence in the correct code. Let $p(y_{\text{true}}|x)$ represent the ALIGN code confidence score (i.e. from the softmax output) for the true code label $y_{\text{true}}$.

    The nonconformity score μ is defined as $\mu = 1 - p(y_{\text{true}}|x)$.

    Higher nonconformity scores indicate greater uncertainty in the ALIGN's code prediction.

2. **Quantile Calculation**: We then determine a critical non-conformity threshold by computing the ϵ-quantile of these scores from the calibration set. This quantile q̂ represents the threshold at which we can ensure the desired coverage (e.g., 95%). i.e. the quantile such that at least 1− ϵ of the true labels in the calibration set fall below the threshold.

3. **Prediction Set Generation**: At test or deployment time, the model computes the confidence scores for a test query $(x_{\text{test}})$. The prediction set $C(x_{\text{test}})$ is constructed by including all predicted candidate codes $y_i$ where the confidence score exceeds the threshold $1-\hat{q}$

   **i**.e. $C(x_{\text{test}}) = \{y_i | \ p(y_i|x_{\text{test}}) \geq 1-\hat{q} \}$

   This provides prediction set adaptivity --- where if ALIGN is confident the set will be small (and even singleton) or if ALIGN is unconfident the set will be larger.

Thus, assuming we have access to labelled data, ALIGN can easily be integrated based on our confidence formulation, offering robust uncertainty estimates and allowing for trustworthy decision-making with frequentist guarantees, improving ALIGN's reliability.

# ALIGN Insights Ablation

To understand the mechanisms behind ALIGN's superior performance over RAG-based methods for ATC coding, we analyzed several components of our system (see Table S4).

- Source of correct code assignments: we found the base LLM was the primary contributor, generating 82.95% and 80.97% of correct codes for RA and SLE respectively. However, the other candidate proposers (BM25 and Dense) contributed non-negligibly, highlighting the value of our hybrid approach in capturing correct codes that the base LLM alone might overlook.
- Self-evaluation filtering mechanism: A significant portion of proposed codes was pruned/filtered by our self-evaluation mechanism (31-33% for LLM, 48-53% for BM25, and 31-33% for Dense) due to entailment failures, underscoring the critical role of our self-evaluation mechanism to improve coding precision.
- Candidate diversity & overlap with LLM: the relatively low overlap between the LLM and retrieval candidates (32-40% for both BM25 and RAG across RA and SLE) demonstrates that the LLM generates diverse candidates, effectively complementing the retrieval-based methods.

*Table S4. Insights into different components of ALIGN. (a) Source of correct code shows the LLM correctly proposes the correct candidate, (b) % of code filtered shows our self-evaluation filters a significant proportion of proposed codes, (c) the low overlap with the LLM shows the need for diverse candidate proposal*

| (a) Correct code source | RA | SLE |
|---|---|---|
| LLM | 82.95 | 80.97 |
| BM25 | 2.58 | 2.60 |
| Dense | 14.47 | 16.42 |
| (b) % candidates filtered/pruned | | |
| LLM | 31 | 32 |
| BM25 | 48 | 53 |
| Dense | 31 | 33 |
| (c) % of candidate overlap w/ LLM | | |
| BM25 | 40 | 36 |
| Dense | 36 | 32 |

# Prompt details.

*Prompt S1. Contextualizer LLM*

> You are an ATC Matching Expert. Your goal is given a query to provide context of the drug. Format the output as follows:
> (1) what it's used for
> (2) what it does
> (3) ingredients
> (4) how it is used (orally, topical, intravenous etc).
>
> ---
>
> Follow the following format.
>
> Query: Input query for which we want to find the ATC code
>
> Context: Description and context of the role of the query medication
>
> ---
>
> Query: {medication name}
>
> Please provide the output field Context. Do so immediately, without additional content before or after, and precisely as the format above shows. Begin with only the field Context.

*Prompt S2. Dense or BM25 candidates--- after retrieval via Dense retrieval or BM25.*

> You are an ATC Matching Expert. Your goal is given a query, context and retrieval info to predict the correct ATC code.
>
> ---
>
> Follow the following format.
>
> Query: Input query for which we want to find the ATC code
>
> Context: Description and context of the query
>
> Context: Retrieved info based on the query on the ATC docs
>
> ATC Codes: All likely predicted ATC code given query, context and retrieval info. Return only the codes (can be more than one).
>
> ---
>
> Query: {medication name}
>
> Context: {Medication context}
>
> Context: {candidates retrieved via Dense retrieval or BM25}
>
> Please provide the output field ATC Codes. Do so immediately, without additional content before or after, and precisely as the format above shows. Begin with only the field ATC Codes.

*Prompt S3. LLM reasoning candidates.*

> You are an ATC Matching Expert. Your goal is given a query, context and alternative names to predict all the likely ATC codes.
>
> ---
> Follow the following format.
>
> Query: {medication name}
>
> Context: {Medication context}
>
> medication info: {dosage and administration route}
>
> ATC Codes: All likely predicted ATC codes given query, context and retrieval info. Return only the codes (can be more than one code) in a comma separated list.

*Prompt S4. Self-evaluation of codes via entailment verification*

> You are an ATC Matching Expert. Given the premise/context, is the hypothesis/query supported or related within the current context and why? Return 1 if supported or 0 if not.
>
> Input hypothesis or query: {query medication}
>
> Additional context about the query: {context on query}
>
> premise: {premise or context on code}
>
> Output
> Supported: {predicted 1/0}
>
> explanation: {reasoning}

*Prompt S5. LLM candidate reasoning for the MCQ predictor*

> Given the concomitant medication query: {query} and context {drug_info}.
>
> Please select the best ATC matches from {candidates after self-evaluation}.
>
> Think step by step and give your reasoning.
>
> Output: {llm_reasoning_context}

*Prompt S6. LLM confidence scoring (extract log probs on output)*

> Which is the correct ATC code for {medication_name} ({medication_info}):
> A ) code 1
> B ) code 2
> C ) code 3
> D ) code 4
> E ) code 5
> F ) None
>
> Which option is correct accounting for the context: {llm_reasoning_context}
>
> Answer with a single letter.